\documentclass[conference]{IEEEtran}
\IEEEoverridecommandlockouts

\usepackage{cite}
\usepackage{amsmath,amssymb,amsfonts}
\usepackage{graphicx}
\usepackage{textcomp}
\usepackage{xcolor}
\usepackage[dvipsnames]{xcolor}
\def\BibTeX{{\rm B\kern-.05em{\sc i\kern-.025em b}\kern-.08em
    T\kern-.1667em\lower.7ex\hbox{E}\kern-.125emX}}

\usepackage[normalem]{ulem}
\usepackage{comment}
\usepackage{algorithm}
\usepackage{algpseudocode}
\algrenewcommand\algorithmicrequire{\textbf{Input:}}
\algrenewcommand\algorithmicensure{\textbf{Output:}}
\usepackage{hhline}
\usepackage{tabularx}
\usepackage{multirow, makecell}
\usepackage{array}

\usepackage[caption=false,font=footnotesize]{subfig}
\usepackage{tikz,amsmath, amssymb,bm,color}
\usetikzlibrary{shapes,arrows}
\usetikzlibrary{calc}
\usetikzlibrary{positioning}
\usetikzlibrary{math}
\usetikzlibrary{arrows.meta}

\makeatletter 
\newcommand{\linebreakand}{%
  \end{@IEEEauthorhalign}
  \hfill\mbox{}\par
  \mbox{}\hfill\begin{@IEEEauthorhalign}
}
\makeatother 

\begin{document}

\title{Adaptive PD Gains for Energy-Conscious Control in Physical Human-Robot Interaction\\
\thanks{The authors would like to acknowledge the support from the University of Calgary, the University of Waterloo RoboHub, as well has the help provided by Alexander Werner and Robert Wagner when carrying out hardware experiments.}
}

\author{\IEEEauthorblockN{Danyal Saqib}
\IEEEauthorblockA{\textit{Department of Electrical and Computer Engineering} \\
\textit{University of British Columbia}\\
Vancouver, Canada \\
dsaqib@student.ubc.ca}
\and
\IEEEauthorblockN{Francisco Andrade Chavez}
\IEEEauthorblockA{\textit{Department of Engineering} \\
\textit{Thompson Rivers University}\\
Kamloops, Canada \\
fandradechavez@tru.ca}
\and

\linebreakand 

\IEEEauthorblockN{Marie Charbonneau}
\IEEEauthorblockA{\textit{Department of Mechanical and Manufacturing Engineering} \\
\textit{University of Calgary}\\
Calgary, Canada \\
marie.charbonneau@ucalgary.ca}
}

\maketitle

\begin{abstract}
Compliant force or torque control are approaches often investigated to achieve safe physical human-robot interaction (pHRI). However, these approaches have limitations. Force control requires a robot to be equipped with external force sensors to track the amplitude and direction of applied forces. Torque control requires torque sensing or estimation in each joint. As this is not available on every robot, energy-based approaches offer a promising alternative. Such approaches aim to achieve safe pHRI by limiting the mechanical energy of the robot. Current schemes leveraging an energy-based approach tend to have a complex implementation, and some may require further stability verification. We hence propose an adaptive proportional-derivative (PD) controller that can limit a robot's energy under any given limit to achieve safe pHRI. The proposed controller can limit both the kinetic and potential energy of a robot, and the behaviour of the controller gains can be shaped using various parameters, defining precisely the cutoff limit and sharpness. We construct a stability proof for the controller and define a condition to ensure the controller's stability. The proposed controller's behaviour and compliance are tested on the TALOS robot from PAL Robotics both in simulation and on hardware, verifying the expected compliant and energy-limiting behaviour of the controller.
\end{abstract}


\section{Introduction} \label{sec: introduction}
Human-Robot Interaction (HRI) is a vast field, aiming to provide safe and meaningful ways for humans and robots to interact. Physical Human-Robot Interaction (pHRI) is a subfield including the study of physical contact between robots and humans in a variety of settings~\cite{pHRI_1, farajtabar_hri}. Safety in pHRI is of utmost importance, being a necessity for HRI applications to find wider applicability. Collisions between a robot and a human may risk a high transfer of energy, which could result in serious human injury. One of the most common ways of ensuring safety for pHRI from a controls perspective is through compliant control. Compliance within this context is defined as control strategies that result in compliant motion - shaping the mechanical impedance or admittance of the robot, aiming to model the robot's response to external forces as a spring-mass-damper system \cite{compliance_1,compliance_2,compliance_3}. However, admittance and impedance control schemes: (i) usually require force or torque sensing to accurately define the relationship between external forces and the robot's position/velocity, (ii) are typically coordinate-dependent, requiring knowledge of applied force direction, and (iii) are often associated with a complicated safety certification process, given that they generally do not provide explicit safety guarantees such as a limit on applied forces as per the ISO/TS 15066 standard~\cite{energy_budget_1, iso_ts_15066_2016}.

In this paper, we propose an adaptive proportional-derivative (PD) controller that can limit a robotic system's energy under a given limit\footnote{Note that the energy limit may be dynamically allocated.}, consequently limiting energy transfer upon contact with the environment. The controller does so by adapting the controller gains depending on how close the current energy of the system is to its specified energy limit. The PD gains only adapt when the energy approaches the limit, and stay close to the nominally tuned values of a traditional PD controller otherwise. The controller is designed to (i) limit the potential energy to reduce clamping dangers, and (ii) limit the kinetic energy to dampen the effects of high external forces and prevent unstable `flailing' movements of the robot. A stability analysis is conducted based on Lyapunov stability theory, yielding a stability condition. Finally, the proposed controller is evaluated on the PAL Robotics TALOS robot, verifying its effectiveness in limiting the system energy and maintaining a compliant behaviour in response to external forces. As PD controllers have been extensively studied, are easy to implement, and can benefit from traditional tuning techniques \cite{kellypd1997}, we expect that the proposed controller may ease the adoption of energy-based control in practical pHRI applications. The introduction of two novel dissipative functions enables limitation of both the controlled robot's energy and the energy that may be injected into the system through interaction with a person, and the tuning parameters allow greater control over the behaviour of the PD gains.
\section{Related Work} \label{sec: related_work}
Towards ensuring safe pHRI, the current version of the ISO/TS 15066 standard applies to collaborative robots and scenarios where humans and robots work in close proximity, specifying maximum permissible velocity, force, pressure, and energy transferred during human-robot contact for each human body part ~\cite{iso_ts_15066_2016}. 
It also specifies that since a robot moving during a contact scenario is potentially dangerous, only a `stop' action is considered safe. However, this policy is unsuitable for collaborative work in tight spaces \cite{energy_budget_1}. In the case of a quasi-static contact scenario (e.g., a robot clamping a human into a wall), a human may not have the means to escape the situation even after the stop action has occurred~\cite{contact_1}. To guarantee human safety during collisions, power and force limiting (PFL) control strategies are often used~\cite{iso_cbf,pfl_1,pfl_2}. They work by either directly reducing the robot's maximum applied forces and velocities during operation, or by detecting collisions and implementing safety measures when applied forces exceed given safety limits. These techniques often require sensing or estimation~\cite{force_estimation_1, force_estimation_2} and a collision reaction scheme since, as mentioned above, a simple stop reaction in response to a collision might not necessarily be safe~\cite{energy_budget_1,collision_reaction_1}.

Energy-limiting control for safe pHRI is a recent area of research promising to address these limitations, given that energy-based safety constraints are generally less conservative than force or velocity-based ones~\cite{iso_1,energy_transfer_1}. As detailed in~\cite{energy_budget_1, iso_1}, energy-limiting approaches can lead to safe robot behaviour as specified in the ISO/TS 15066 standards. Energy as a quantity is also coordinate-independent, thus making energy-based safety constraints more convenient in situations where contact force estimation depends on the robot configuration \cite{force_direction_1, iso_1, energy_budget_1}.  
The authors of~\cite{energy_tank_1} utilize the concept of `virtual energy tanks' to ensure the passivity of an impedance controlled robot with time-varying stiffness---they do so by tracking the energy dissipated due to damping, and ensuring the energy introduced due to varying stiffness is always less than the energy dissipated. This concept is utilized in~\cite{iso_1} to limit a robot's kinetic energy at all times. As a result, the energy transferred by a robot to a human during a collision is kept under a maximum allowable limit. A different approach is proposed in~\cite{energy_budget_1}, where the stiffness parameters of an impedance controller are modified if the total system energy violates a given energy limit. An increase in stiffness is also utilized to oppose external pushes that induce excessive kinetic energy. Energy-based safety indicators are determined in \cite{meguenani2017energy}, based on human-robot proximity and maximum allowable forces, then used as constraints in an optimization-based controller. An energy-limiting control scheme via changes to the reference position is proposed in \cite{laffranchi2009safe} for a single Series Elastic Actuation (SEA) joint. 
An alternative control scheme capable of limiting robot power and energy using stiffness scaling and damping injection is presented in \cite{hjorth2020energy}, where artificial repulsive potential fields are utilized for cartesian constraints and joint limit avoidance. The authors of \cite{choi2024multi} build upon the work in \cite{hjorth2020energy} by extending energy-limiting control to multi-task operations for redundant robots. The authors of both \cite{hjorth2020energy} and \cite{choi2024multi} establish stability through power regulation. A passivity-based energy-limiting controller is presented in \cite{groothuis2018stabilityenergy}, with a budget allocation algorithm that allocates energy budgets to each individual actuator. However, this approach is yet to be validated for pHRI. \cite{folkertsma2018safetystability} presents a method to guarantee passivity for the actuators of a robot in case communication is lost with higher-level controllers. \cite{zanella2024optimal,califano2024effectcontrolbarrierfunctions} both analyse port-Hamiltonian systems and propose energy-based control strategies, but lack implementations and experimental validations. Overall, the control algorithms introduced above may be complex to integrate into existing robotic control schemes since works such as \cite{iso_1, hjorth2023energystability, choi2024multi} require implementations of energy tanks alongside the controller to maintain passivity and power regulation, whereas works such as \cite{hjorth2020energy, energy_budget_1} lack explicit stability proofs. Current control schemes are also reactive rather than proactive---the torque is modified only when the robot's energy exceeds the budget. These factors motivate the work presented below, integrating energy-based control into a simple, familiar and proven PD control scheme that is also proactive.

The rest of the paper is organized as follows: Sec.~\ref{sec: background} presents some background for robot energy and PD control. Sec.~\ref{sec: energy limiting adaptive pd gains} develops the proposed controller equations for energy-limitation. In Sec.~\ref{sec:stability analysis}, we construct a stability proof of the controller based on the Lyapunov stability theory and provide an analysis of the obtained stability condition. Sec.~\ref{sec: experiments} presents the results for controller validation in simulation and on hardware. Sec.~\ref{sec: discussion} discusses the results, along with the advantages and limitations of the proposed approach. Finally, Sec.~\ref{sec: conclusion} concludes the paper by summarizing our work and discussing possible future directions.


\section{Background} \label{sec: background}
The general form of an $n$-dof fixed-base robot's dynamic model\cite{kelly2005control,spong2005robot} can be expressed as follows :
\begin{align}
\boldsymbol{M}(\boldsymbol{q}) \ddot{\boldsymbol{q}} + \boldsymbol{C}(\boldsymbol{q}, \dot{\boldsymbol{q}})\dot{\boldsymbol{q}} + \boldsymbol{G}(\boldsymbol{q}) = \boldsymbol{\tau} + \boldsymbol{\tau}_{ext} \label{robot_model}
\end{align}
where $\boldsymbol{q}\in\mathbb{R}^n$ is the joint position vector, $\boldsymbol{M}(\boldsymbol{q}) \in\mathbb{R}^{n\times n}$ is the mass/inertia matrix, $\boldsymbol{C}(\boldsymbol{q}, \dot{\boldsymbol{q}})\in\mathbb{R}^{n\times n}$ is the Coriolis/centripetal matrix, $\boldsymbol{G}(\boldsymbol{q})\in\mathbb{R}^n$ is the gravity vector, $\boldsymbol{\tau}\in\mathbb{R}^n$ is the joint torque vector, and $\boldsymbol{\tau}_{ext}\in\mathbb{R}^n$ is the vector of externally applied torques. A notable property of $\boldsymbol{M}$ and $\boldsymbol{C}$ is that $\boldsymbol{\dot{M}}(\boldsymbol{q}) - 2 \boldsymbol{C}(\boldsymbol{q}, \dot{\boldsymbol{q}})$ yields a skew-symmetric matrix. In task-space coordinates, Eq.~\eqref{robot_model} translates to:
\begin{align}
    \boldsymbol{\Lambda} (\boldsymbol{x}) \ddot{\boldsymbol{x}} + \boldsymbol{\mu}(\boldsymbol{x}, \dot{\boldsymbol{x}}) \dot{\boldsymbol{x}} + \boldsymbol{\mathcal{F}}_g(\boldsymbol{x}) = \boldsymbol{\mathcal{F}}_{\tau} + \boldsymbol{\mathcal{F}}_{ext}
\end{align}
where $\boldsymbol{x}\in\mathbb{R}^6$ refers to the end-effector pose (including position and orientation), $ \boldsymbol{\Lambda} (x)$, $\boldsymbol{\mu}(x, \dot{x}) \in\mathbb{R}^{6\times6}$ and $\boldsymbol{\mathcal{F}}_g(\boldsymbol{x})\in\mathbb{R}^6$ are the task-space equivalents of the mass matrix, Coriolis matrix, and gravity vector, respectively. $\boldsymbol{\mathcal{F}}_\tau\in\mathbb{R}^6$ is the end-effector wrench resulting from joint torques and $\boldsymbol{\mathcal{F}}_{ext}\in\mathbb{R}^6$ is the externally applied wrench to the end-effector. The goal of traditional impedance controllers is to shape the mechanical impedance of a robot \cite{ott_cartesian,kelly2005control}, which may be expressed as:

\begin{equation}  \label{desired_impedance}
    \boldsymbol{\Lambda}_d \boldsymbol{\ddot{\tilde{x}}} + \boldsymbol{D}_d \boldsymbol{\dot{\tilde{x}}} + \boldsymbol{K}_d \boldsymbol{\tilde{x}} = \boldsymbol{\mathcal{F}}_{ext} 
\end{equation}
where, $\boldsymbol{\Lambda}_d$, $\boldsymbol{D}_d$, and $\boldsymbol{K}_d$ are the symmetric and positive definite matrices of the desired task-space inertia, damping, and stiffness, respectively, and $\boldsymbol{\tilde{x}} = \boldsymbol{x} - \boldsymbol{x}_r$ given a reference position $\boldsymbol{x}_r$. An appropriate function $\boldsymbol{\mathcal{F}}_{\tau}$ can then be chosen to achieve a desired impedance. 

Physical interaction between a robot and its environment can be described as an energy flow between the two \cite{energy_budget_1}. The energy of a controlled robot can be defined in the form:
\begin{equation}  \label{eq:energy}
    \mathcal{L} = T(\dot{\boldsymbol{q}}, \boldsymbol{M}(\boldsymbol{q})) + 
    U_q(\boldsymbol{\tilde{q}}, \boldsymbol{K}_{q})
\end{equation}
where $T(\dot{\boldsymbol{q}}, \boldsymbol{M}(\boldsymbol{q}))$ is the kinetic energy, and $U_{q}(\boldsymbol{\tilde{q}}, \boldsymbol{K}_{\boldsymbol{q}})$ is the controlled joint potential energy, with $\boldsymbol{\tilde{q}} = \boldsymbol{q} - \boldsymbol{q}_r$ given a reference joint position $\boldsymbol{q}_r$, and $\boldsymbol{K}_{\boldsymbol{q}}$ the joint stiffness matrix. 
A $U_g$ term may be added in Eq.~\eqref{eq:energy} to include the gravitational potential energy. However, it may be neglected under the assumption that gravity is compensated by the controller. In the case of pHRI, gravity compensation is critical to avoid the robot collapsing~\cite{energy_budget_1}, and to prove stability~\cite{spong2005robot,kelly2005control}. The equation of a gravity-compensated PD controller can be given as:
\begin{equation}
    \boldsymbol{\tau} = - \boldsymbol{K}_p \boldsymbol{\tilde{q}} - \boldsymbol{K}_d \boldsymbol{\dot{\tilde{q}}} + \boldsymbol{G}(\boldsymbol{q}) \label{pd_grav}
\end{equation}
where the diagonal proportional and derivative gain matrices $\boldsymbol{K}_p$ and $\boldsymbol{K}_d \in \mathbb{R}^{n \times n}$  correspond to the stiffness and damping of the controlled joints. Note that they may be constant, or adaptive---in which case they become functions of time.

Following the ISO/TS 15066 standard, the maximum allowable energy transfer $\mathcal{L}_{max}$ between a robot and a human body part can be calculated using the maximum permissible force $f_{max}$ and the stiffness or related effective spring constant $k$ for this body part, using the following equation:
\begin{align}
    \mathcal{L}_{max} = \frac{f_{max}^2}{2k} \label{max_energy_calculation}
\end{align}
Other methods of calculating the energy transferred can be found in the literature, such as in~\cite{energy_transfer_1}.
\begin{figure*}[!t]
    \centering
    \def\twidth{0.1}
    \input{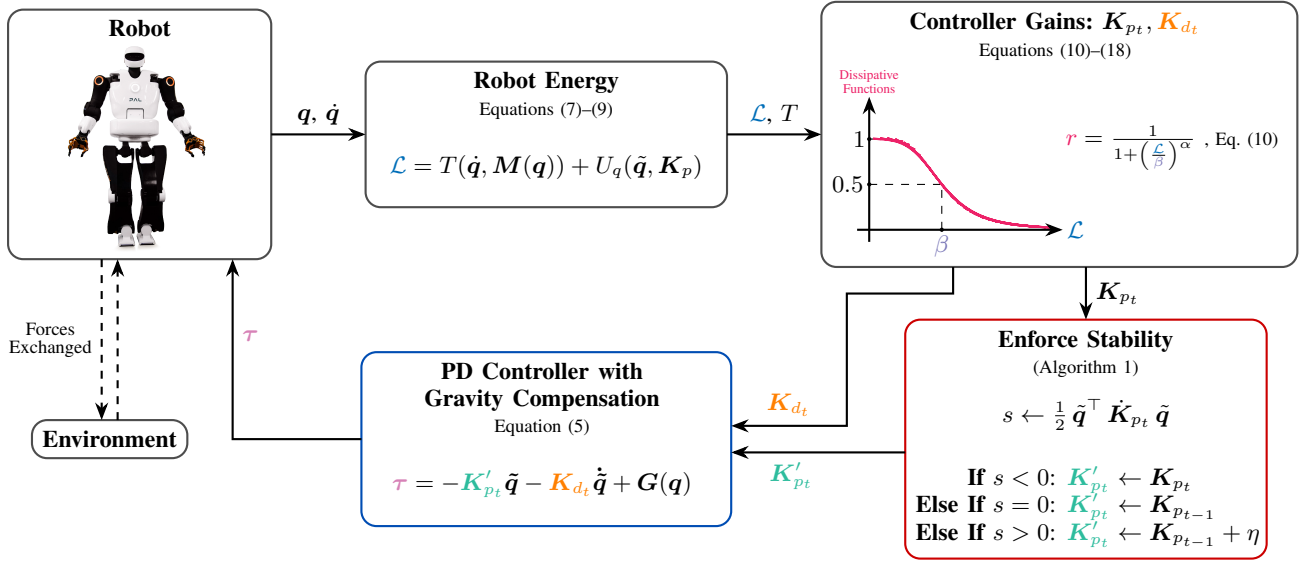}
    \caption{An overview of the proposed energy-limiting control approach. The robot's energy at each instant is used to to determine the value of the dissipative functions ($r$, $r_T$), which in turn determine the PD controller gains. Stability is checked and enforced by modifying $\boldsymbol K_{p_t}$ if necessary. These PD gains alongside gravity compensation determine the resulting joint torques applied by the actuators.
    }
    \label{box_diagram}
\end{figure*}
\section{Energy-Limiting Adaptive PD Gains} \label{sec: energy limiting adaptive pd gains}

This section describes our proposed PD gain adaptation approach, such that the energy of a robot subjected to the control law of Eq.~\eqref{pd_grav} remains below a specified energy limit. The main idea is to introduce two dissipative functions for scaling the PD gains: $r$ is a function of energy that has a value of close to 1 for values of energy well below the energy budget, and tends to zero (or dissipates) as we approach the energy limit. This function can be used to scale the proportional gains of the controller; proportional gains are scaled down as we approach the energy limit. The dissipative function $r_T$ is used to add damping to the system in case of excessive kinetic energy. Both of these functions are discussed in further detail in subsection~\ref{sub:dissipative terms}.

\subsection{System Energy}
The kinetic, potential and total energy of a PD controlled robot may be defined as follows:
\begin{align}
T &= \frac{1}{2} \; \dot{\boldsymbol{q}}^\top \; \boldsymbol{M} (\boldsymbol{q}) \; \dot{\boldsymbol{q}} \label{kinetic_energy} \\
U_q &= \frac{1}{2} \; \boldsymbol{\tilde{q}}^\top \; \boldsymbol{K}_p \; \boldsymbol{\tilde{q}} \label{potential_energy} \\
\mathcal{L} &= T + U_q \label{robot_energy}
\end{align}

\subsection{Nominal PD Gains}
Let us suppose that the PD gains are tuned using any tuning method, then stored in diagonal matrices $\boldsymbol{K}_{{po}_{diag}}$ and $\boldsymbol{K}_{{do}_{diag}} \in \mathbb{R}^{n \times n}$. 
These `nominal' gains will serve as the desired values for when the system energy is well below the given limit. 
We also define the following two operations for forward and inverse diagonalization:
\begin{align*}
\boldsymbol{A}  = 
\begin{bmatrix}
a_{11}\\
a_{22}\\
.. \\
a_{nn}
\end{bmatrix}, \quad
\boldsymbol{B} = 
\begin{bmatrix}
a_{11} & 0 & .. & 0\\
0 & a_{22} & .. & 0\\
0 & 0 & .. & a_{nn}
\end{bmatrix}
\end{align*}
\begin{align*}
diag(\boldsymbol{A}) = \boldsymbol{B} , \quad
diag^{-1}(\boldsymbol{B}) = \boldsymbol{A} 
\end{align*}
yielding column vectors for the nominal PD gains:
\begin{align*}
    \boldsymbol{K}_{po} = diag^{-1}(\boldsymbol{K}_{{po}_{diag}}) \\
    \boldsymbol{K}_{do} = diag^{-1}(\boldsymbol{K}_{{do}_{diag}})
\end{align*}

\subsection{The dissipative functions $r$ and $r_{T}$} \label{sub:dissipative terms}
Let us suppose there exists an upper limit on the system energy, denoted by $\mathcal{L}_{lim}$. Our first contribution is to define a dissipative function $r$ as follows:
\begin{align}\label{r_function}
r = \frac{1}{1 + \left(\frac{\mathcal{L}}{\beta}\right)^\alpha}
\end{align}
where the parameter $\beta < \mathcal{L}_{lim}$ is any cutoff energy point close to which the PD gains start to change, and the parameter $\alpha > 1$ defines how sharp the cutoff is. The value of $r$ is close to 1 when the system energy $\mathcal{L}$ is low, and tends toward 0 as $\mathcal{L}$ increases. Multiplying the nominal PD gains with $r$ will achieve a dissipative behaviour, yielding gain values that remain close to the nominal value when $\mathcal{L}$ is low, and that tend toward 0 as $\mathcal{L} \rightarrow \mathcal{L}_{lim}$. 

We introduce a second dissipative term $r_{T}$ as follows:
\begin{align}
    r_{T} &= \frac{1}{1 + \left(\frac{T}{\beta}\right)^\alpha} \label{r_ke_function}
\end{align}
This term is used to address the problem of user-added energy. When a robot physically interacts with a human, there is a two-way energy transfer that takes place between them~\cite{iso_1,energy_budget_1}. 
The human may transfer energy to the robot, for example when pushing it away during a clamping scenario. This would cause the robot to deviate from a desired trajectory, leading to increased $\dot{\boldsymbol{q}}$ and $\boldsymbol{\tilde{q}}$. The $r_T$ function is designed to return a value close to 0 when the system's kinetic energy is low, and approaching 1 as the kinetic energy approaches $\mathcal{L}_{lim}$. This term can hence be used to increase the damping of the robot in response to high kinetic energy. Its implementation is discussed in further detail in Sec.~\ref{ref: subsec kd}.
\subsection{The Adaptive Proportional Gain Matrix $\boldsymbol{K}_p$}
The adaptive proportional gain matrix $\boldsymbol{K}_p \in \mathbb{R}^{n\times n}$ for our proposed controller is defined as:
\begin{align}
    \boldsymbol{K}_{p} = diag(\boldsymbol{p}_1^\top \boldsymbol{p}_2 \label{K_p}) 
\end{align}
where the terms $\boldsymbol{p}_1$ and $\boldsymbol{p}_2$ are obtained using the following:
\begin{align}
    \boldsymbol{p}_1 &=  r \boldsymbol{K}_{po} + (1 - r) \frac{\mathcal{L}_{lim}}{n} \label{p1} \\
    \boldsymbol{p}_2 &= 1 \oslash \left[1 + (1 - r)  \boldsymbol{\tilde{q}}^2 \right] \label{p2} 
\end{align}
where $\oslash$ denotes the Hadamard division operator. In Eq.~\eqref{p1}, $r \boldsymbol{K}_{po}$ results in the proportional gains being dissipated as $\mathcal{L}$ increases, and $(1 - r) \frac{\mathcal{L}_{lim}}{n}$ ensures that as $\mathcal{L} \rightarrow \mathcal{L}_{lim}$, $\boldsymbol{p}_1 \rightarrow \frac{\mathcal{L}_{lim}}{n}$, dividing the total energy budget equally among all of the joints. This design choice was made for simplicity. Other strategies for dividing the total budget exist, such as using desired joint positions and velocities for actuator budgets \cite{groothuis2018stabilityenergy}, or inertia-based budget allocation. The focus of our research was energy limitation given a budget rather than optimal allocation of the budget. The term $\boldsymbol{p}_2$ is a dissipative term, designed to keep $\boldsymbol{K}_p$ as high as possible without violating $\mathcal{L}_{lim}$, such that when $\mathcal{L} \rightarrow \mathcal{L}_{lim}$, $\boldsymbol{K}_p \rightarrow diag \left(\frac{\mathcal{L}_{lim}}{n} \oslash (1 + \boldsymbol{\tilde{q}}^2)\right)$. As the energy limit approaches, a lower value of $\boldsymbol{K}_{p}$ causes a direct reduction in potential energy.

\subsection{The Derivative Gain Matrix $\boldsymbol{K}_d$} \label{ref: subsec kd}
The adaptive derivative gain matrix $\boldsymbol{K}_d$ for our proposed controller is defined as:
\begin{align}
    \boldsymbol{K}_d &= diag\left( \boldsymbol{d}_1^\top \boldsymbol{d}_2 + \boldsymbol{d}_3 \right)
\end{align}
where the terms $\boldsymbol{d}_1$, $\boldsymbol{d}_2$ and $\boldsymbol{d}_3$ are defined as follows:
\begin{align}
    \boldsymbol{d}_1 &= r \boldsymbol{K}_{do} \label{d1} \\
    \boldsymbol{d}_2 &= 1 \oslash \left[1 + (1 - r) \dot{\boldsymbol{q}}^2 \right]\label{d2} \\
    \boldsymbol{d}_3 &= (1 - r_{T}) \boldsymbol{K}_{d_{max}} \label{d3}
\end{align}
$\boldsymbol{K}_{d_{max}} \in \mathbb{R}^{n}$ is a tunable parameter defining an upper limit to $\boldsymbol{K}_{d}$.
The term $\boldsymbol{d}_1$ results in $\boldsymbol{K}_d$ being dissipated as $\mathcal{L}$ rises. The term $\boldsymbol{d}_2$ is a dissipative term working in a similar way as $\boldsymbol{p}_2$. The term $\boldsymbol{d}_3$ addresses the problem of user-added energy by opposing potentially uncontrollable movements of the robot in response to external pushes, as described in subsection~\ref{sub:dissipative terms}. If $\mathcal{L}_{lim}$ is approached in this manner, $\boldsymbol{K}_p$ and $\boldsymbol{K}_d$ will tend to decrease, leading to the robot offering little to no resistance to a push. Hence, the user-added energy will lead to a possible violation of the energy limit. To deal with this issue, we identify one key factor: if the energy limit is violated due to an external push to the robot, this will be reflected almost entirely in the kinetic energy of the system. This is because if the robot is pushed away and the energy increases, both the PD gains will adaptively decrease, leading to a direct decrease in the potential energy, as defined in Eq.~\eqref{potential_energy}. The energy violation hence must be a consequence of high kinetic energy. Increasing the derivative gain $\boldsymbol{K}_d$ as the kinetic energy approaches $\mathcal{L}_{lim}$ can offer a way of resisting the push, without violating the energy limit. This is equivalent to increasing the damping in case of high external forces, to resist a potentially uncontrollable motion of the robot. Note that it is important that the opposition to user-added energy be in the form of increased damping, to prevent high joint velocities and high reactive forces. If the opposition was instead in the form of increased elastic potential, the controller would respond with an undesirable reactive force to reduce tracking error~\cite{energy_budget_1}. To provide adequate opposition, $\boldsymbol{K}_{d_{max}}$ must be carefully tuned: an arbitrarily high value could result in undesirably high resistance to external forces. However, during a clamping scenario, we want the human to be able to push the robot away.

\section{Stability Analysis} \label{sec:stability analysis}
In this section, we provide a stability analysis for our proposed controller, based on Lyapunov stability theory applied to PD controllers~\cite{kelly2005control,passivity_2}. We obtain a condition for the stability of the controlled system, and propose an algorithmic way of
incorporating it into the controller.

A common assumption in PD control stability proofs is to have a constant reference position $\boldsymbol{q}_r$, such that $\boldsymbol{\dot{\tilde{q}}} = \dot{\boldsymbol{q}}$~\cite{kelly2005control}. By making this assumption, the stability of the controller is proven for a constant position tracking problem, but it can be shown that if $\boldsymbol{q}_r(t)$, then $\boldsymbol{\dot{\tilde{q}}}$ becomes asymptotically small with sufficiently large gains~\cite{kelly2005control,pd_stability_q_d}. In our approach, the gains are generally able to increase as long as a sufficient energy budget is available. 
Cases where energy budgets are too low may potentially lead to low gains. However, perpetually low gains will affect convergence to the reference position rather than leading to unstable behaviour---the system may not reach the desired position, but unstable behaviour or oscillations are not expected to occur due to low gains alone.

\subsection{The Lyapunov Analysis}
The closed loop equation for a robot controlled with the proposed controller is given by substituting Eq.~\eqref{pd_grav} into Eq.~\eqref{robot_model}, yielding:
\begin{align} \label{eq:closed_loop_equation}
\boldsymbol{M}(\boldsymbol{q}) \; \ddot{\boldsymbol{q}} &= - \boldsymbol{K}_p (t) \; \boldsymbol{\tilde{q}} - \boldsymbol{K}_d (t) \; \boldsymbol{\dot{\tilde{q}}} - \boldsymbol{C}(\boldsymbol{q}, \dot{\boldsymbol{q}}) \; \dot{\boldsymbol{q}}
\end{align}
Solving Eq.~\eqref{eq:closed_loop_equation} for joint accelerations yields:
\begin{align}
    \ddot{\boldsymbol{q}} = \boldsymbol{M}^{-1}(\boldsymbol{q}) \left( - \boldsymbol{K}_p (t) \; \boldsymbol{\tilde{q}} - \boldsymbol{K}_d (t) \; \boldsymbol{\dot{\tilde{q}}} - \boldsymbol{C}(\boldsymbol{q}, \dot{\boldsymbol{q}}) \; \boldsymbol{\dot{\tilde{q}}} \right) \label{joint_acc_eqn}
\end{align}
We propose the following candidate Lyapunov function:
\begin{align}
    V = \frac{1}{2} \boldsymbol{\dot{\tilde{q}}}^\top \; \boldsymbol{M}(\boldsymbol{q}) \; \boldsymbol{\dot{\tilde{q}}} + \frac{1}{2} \; \boldsymbol{\tilde{q}}^\top \; \boldsymbol{K}_{p} (t) \; \boldsymbol{\tilde{q}}
\end{align}
The time derivative of $V$ yields:
\begin{align}
    \dot{V} = {}&\boldsymbol{\dot{\tilde{q}}}^\top \; \boldsymbol{M}(\boldsymbol{q}) \; \boldsymbol{\ddot{\tilde{q}}} + \frac{1}{2} \; \boldsymbol{\dot{\tilde{q}}}^\top \; \boldsymbol{\dot{M}}(\boldsymbol{q}) \; \boldsymbol{\dot{\tilde{q}}} + \; \boldsymbol{\tilde{q}}^\top \; \boldsymbol{K}_{p} (t) \; \boldsymbol{\dot{\tilde{q}}} \nonumber \\
    &{+}\: \frac{1}{2} \; \boldsymbol{\tilde{q}}^\top \; \boldsymbol{\dot{K}}_p (t) \; \boldsymbol{\tilde{q}} \label{lyap_dot}
\end{align}
Substituting Eq.~\eqref{joint_acc_eqn} into Eq.~\eqref{lyap_dot}, and given that the quadratic form of the skew-symmetric matrix $\boldsymbol{\dot{M}}(\boldsymbol{q}) - 2 \boldsymbol{C}(\boldsymbol{q}, \dot{\boldsymbol{q}})$ is zero, we obtain:
\begin{align}
    \dot{V} &= - \boldsymbol{\dot{\tilde{q}}}^\top \boldsymbol{K}_d (t) \; \boldsymbol{\dot{\tilde{q}}} + \frac{1}{2} \; \boldsymbol{\tilde{q}}^\top \; \boldsymbol{\dot{K}}_p (t) \; \boldsymbol{\tilde{q}}
\end{align}
Following the Lyapunov Stability Theorem, $\dot{V}$ must be negative definite to establish stability. For $\dot{V} < 0$ to hold, then the following condition must be true:
\begin{align}
    \boxed{ \boldsymbol{\dot{\tilde{q}}}^\top \boldsymbol{K}_d (t) \; \boldsymbol{\dot{\tilde{q}}} > \frac{1}{2} \; \boldsymbol{\tilde{q}}^\top \; \boldsymbol{\dot{K}}_p (t) \; \boldsymbol{\tilde{q}}} \label{stability_condition}
\end{align}
Section~\ref{sec:stability condition} analyzes the implications of Eq.~\eqref{stability_condition}.
\begin{algorithm}[!t]
\caption{Enforcing the Stability Condition}\label{algorithm_stability_enforcement}
\begin{algorithmic}[1]
	\Require{$\boldsymbol{K}_{{d}_{t}}, \boldsymbol{K}_{{p}_{t}}, \boldsymbol{K}_{{p}_{t-1}}, \eta$} \vskip 3pt
	\Ensure{$\boldsymbol{K}^\prime_{{p_t}}$}\vskip 5pt
	\State $s \gets \frac{1}{2} \; \boldsymbol{\tilde{q}}^\top \; \boldsymbol{\dot{K}}_{p_t} \; \boldsymbol{\tilde{q}}$ \vskip 3pt
	\If{$s < 0$} \vskip 3pt
    		\State $\boldsymbol{K}^\prime_{{p_t}} \gets \boldsymbol{K}_{{p}_{t}}$ \vskip 5pt
	\ElsIf{$s = 0$} \vskip 3pt
    		\State $\boldsymbol{K}^\prime_{{p_t}} \gets \boldsymbol{K}_{{p}_{t-1}}$ \vskip 5pt
        \ElsIf{$s > 0$} \vskip 3pt
    		\State $\boldsymbol{K}^\prime_{{p_t}} \gets \boldsymbol{K}_{{p}_{t-1}} + \eta$ \vskip 3pt
	\EndIf
\end{algorithmic}
\end{algorithm}
\begin{figure*}[!t]
\centering
\def\twidth{0.1}
\subfloat[Traditional PD Controller Sinusoidal Tracking]{\includegraphics[width=3.4in]{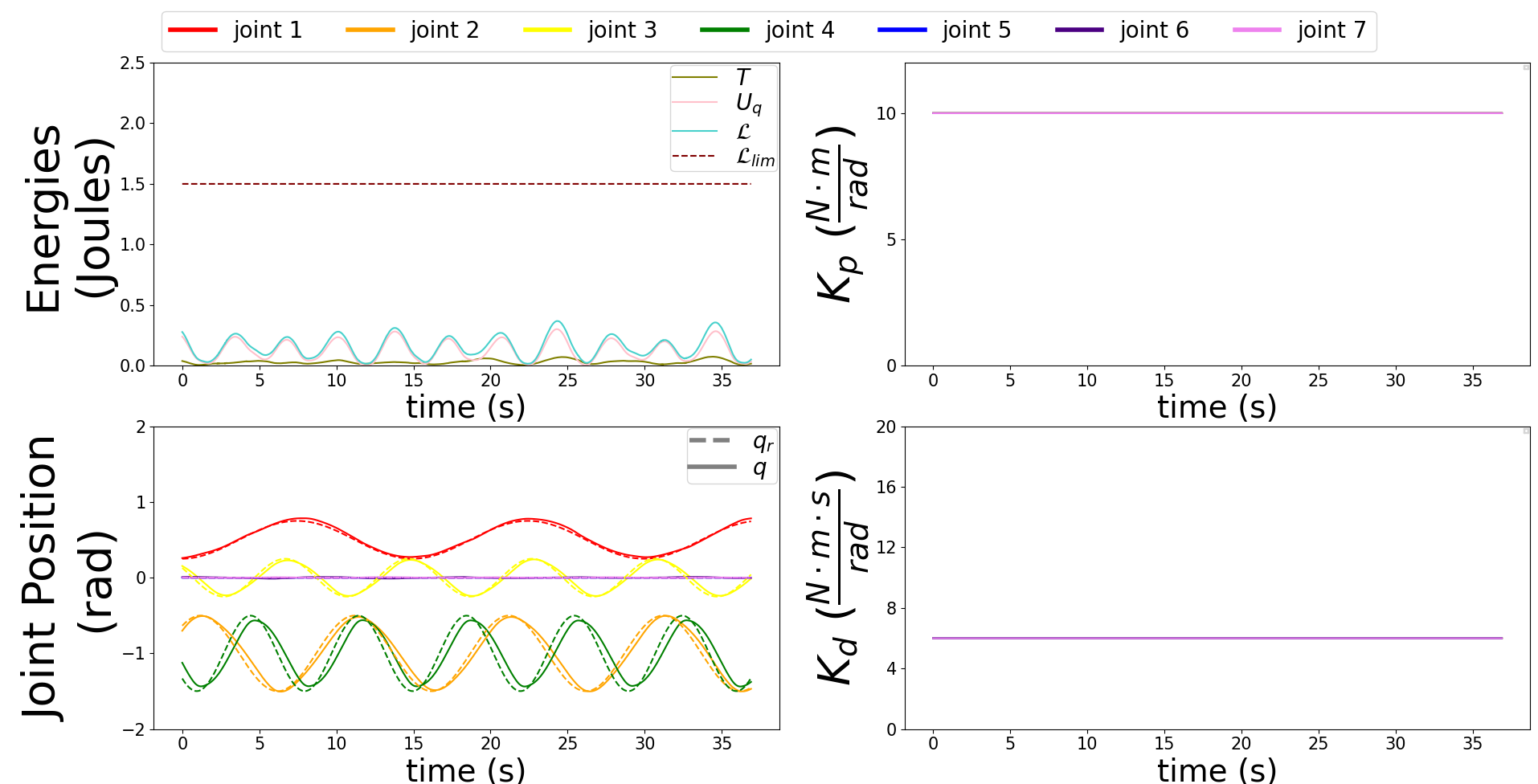}
\label{fig_sine_simple}}
\hfil
\subfloat[Adaptive PD Controller Sinusoidal Tracking]{\includegraphics[width=3.4in]{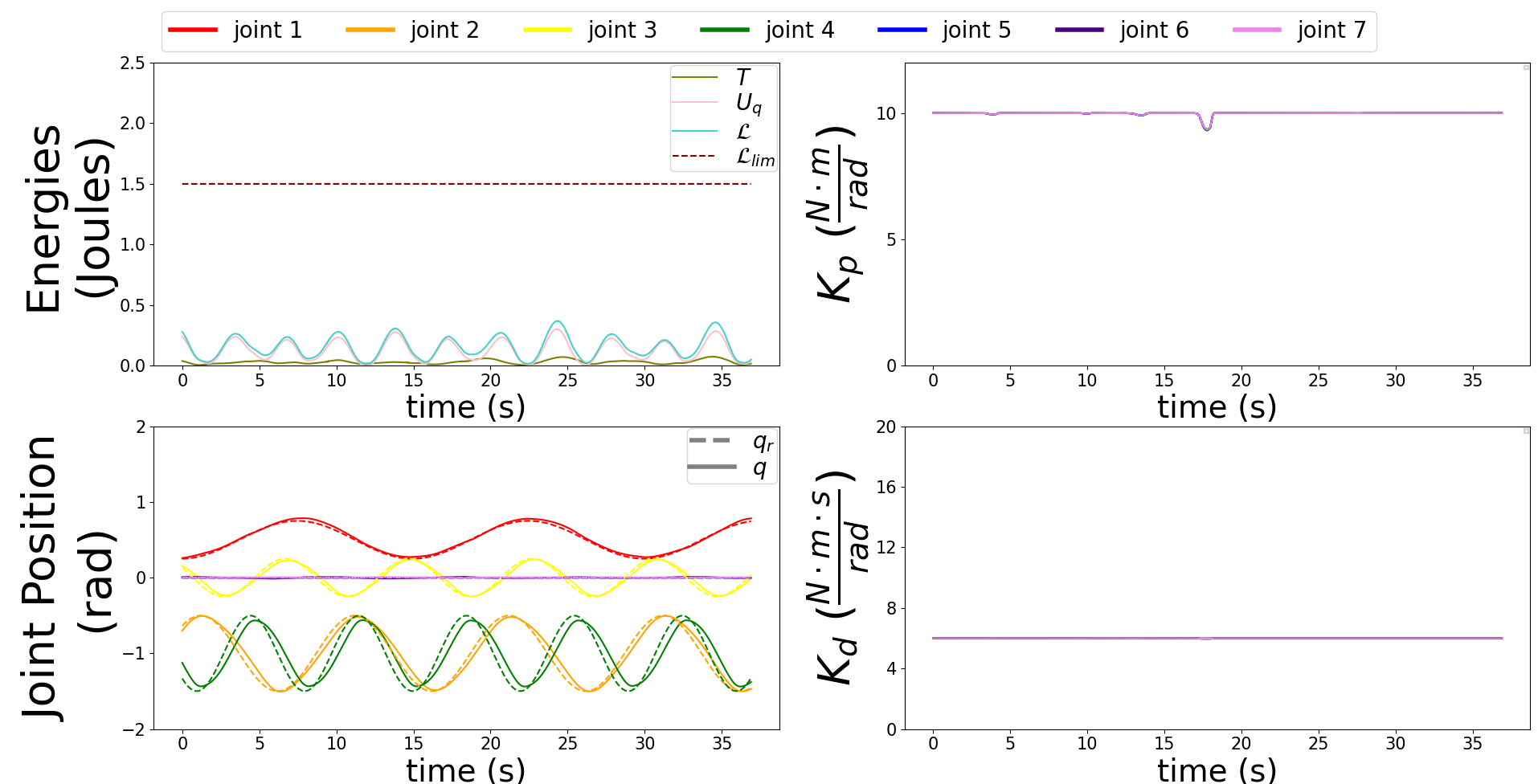}
\label{fig_sine_adaptive}}
\caption{Controller responses in simulation given sinusoidal reference trajectories for joints 1-4 of the right arm; joints 5-7 are given a 0 reference. Reference sinusoidal parameters are found in Tab.~\ref{tab:sinusoid parameters}. Responses are shown for \protect\subref{fig_sine_simple} the traditional non-adaptive PD controller and \protect\subref{fig_sine_adaptive} the proposed adaptive PD controller.}
\label{fig_sine}
\end{figure*}

\begin{figure*}[!t]
    \centering
    \def\twidth{0.1}
    \subfloat[Traditional PD Controller Response]{\includegraphics[width=3.4in]{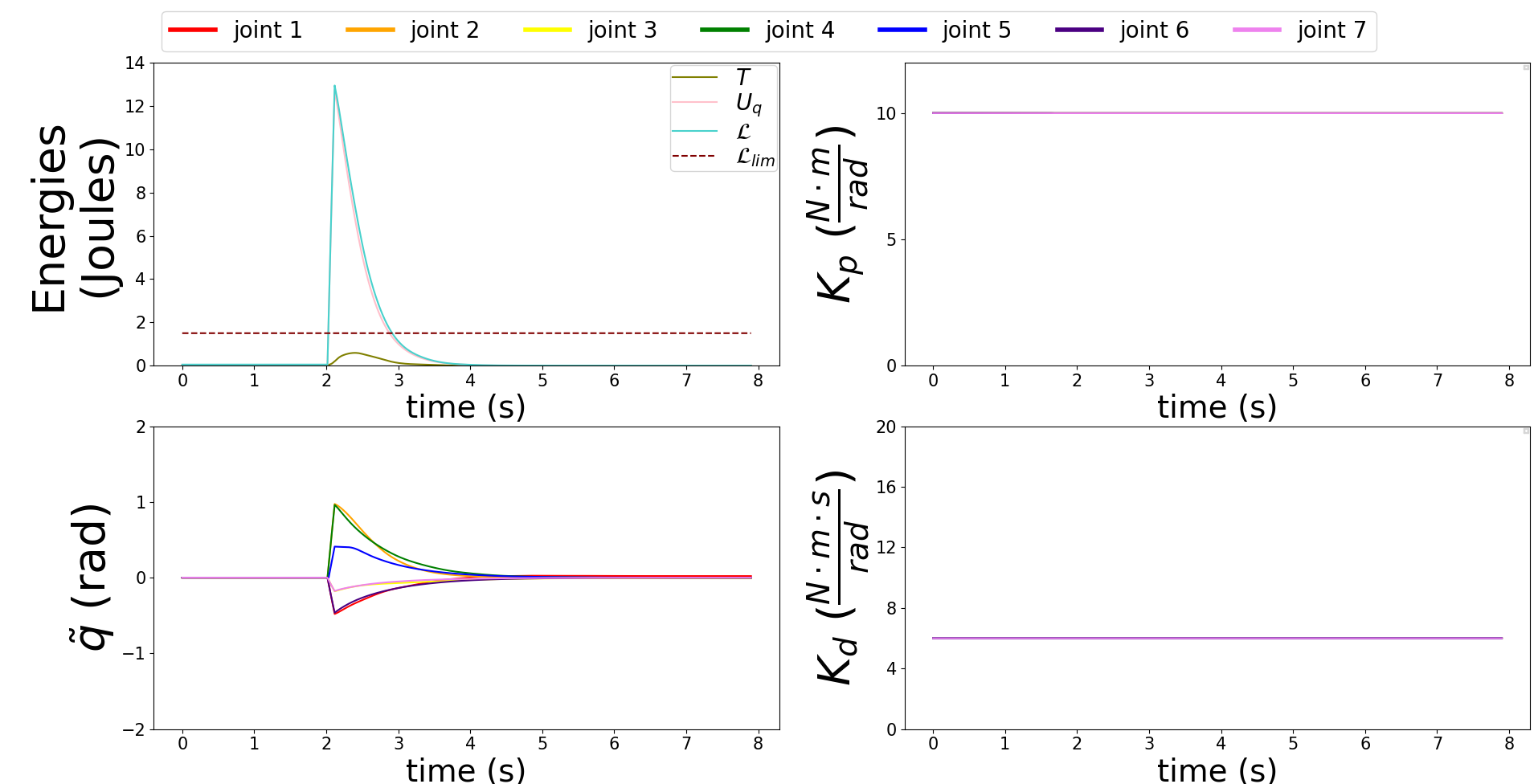}
    \label{fig_tracking_simple}}
    \hfil
    \subfloat[Adaptive PD Controller Response]{\includegraphics[width=3.4in]{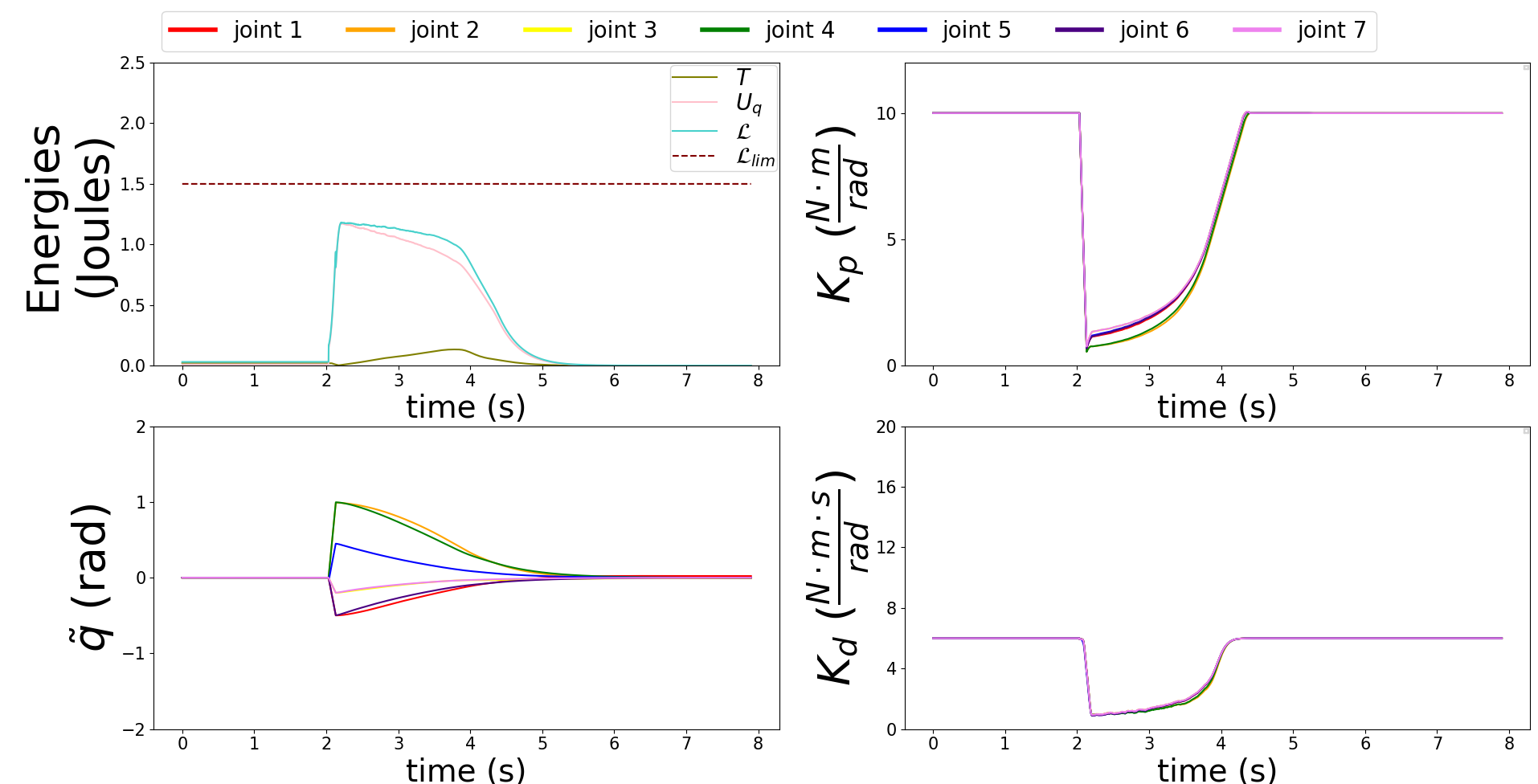}
    \label{fig_tracking_adaptive}}
    \caption{Controller responses in simulation for a reference position step input to $\boldsymbol{q}_r = \begin{bmatrix}
        0.5 & -1 & 0.2 & -1 & -0.5 & 0.5 & 0.2
    \end{bmatrix}^\top~\text{rad}$ occurring at the $2~\text{s}$ mark for \protect\subref{fig_tracking_simple} the traditional non-adaptive PD controller and \protect\subref{fig_tracking_adaptive} the proposed adaptive PD controller. The step input occurs at the $2~\text{s}$ mark. 
    }
    \label{fig_tracking}
\end{figure*}
\subsection{The Stability Condition}\label{sec:stability condition}
For the proposed adaptive PD controller to be stable, the condition of Eq.~\eqref{stability_condition} must remain true. 
Observe that $\boldsymbol{K}_d (t)$ is positive definite. Thus,
\begin{align*}
    \boldsymbol{\dot{\tilde{q}}}^\top \boldsymbol{K}_d (t) \; \boldsymbol{\dot{\tilde{q}}} \geq 0 \quad \forall \quad \boldsymbol{\dot{\tilde{q}}} \in \mathbb{R}^n 
\end{align*}
This implies that when $\boldsymbol{\dot{K}}_p (t) < 0$, there is no limit on the rate of change of $\boldsymbol{K}_p$, such that it may decrease as fast as desired. The condition imposes an upper bound on $\boldsymbol{\dot{K}}_p (t)$, such that $\boldsymbol{K}_p$ may not increase arbitrarily fast. We define Algorithm \ref{algorithm_stability_enforcement} to ensure stability according to Eq.~\eqref{stability_condition}. If $\boldsymbol{K}_p$ is small and there is no change in the error (i.e., $\boldsymbol{\dot{\tilde{q}}}=0$) , according to Eq.~\eqref{stability_condition} the upper bound on $\boldsymbol{\dot{K}}_p (t)$ becomes zero---$\boldsymbol{K}_p$ may never increase and the controlled system's position may not change. This problem is also discussed in~\cite{hi-per}, where the stability of a cartesian impedance controller is analysed and a similar stability condition is obtained. 
However, \cite{hi-per} also observe no unstable or nearly unstable behaviour under step-increased stiffness, despite momentary violations of the stability condition. Hence, our algorithm also opts for a step-increased stiffness approach. It relies on discretizing the rate of change of $\boldsymbol{K}_p (t)$ as:
\begin{align*}
    \boldsymbol{\dot{K}}_{p_t} = \frac{\boldsymbol{K}_{p_t} - \boldsymbol{K}_{{p}_{t-1}}}{dt}
\end{align*}
where $dt$ is the controller sample time. 
The $t$ and $t-1$ subscripts denote the current and previous timesteps, respectively. Used in Algorithm~\ref{algorithm_stability_enforcement}, the tunable `step size' parameter $\eta$ determines how fast the value of $\boldsymbol{K}_p (t)$ may rise.
The $\boldsymbol{K}^\prime_{{p}_{t}}$ term output by Algorithm~\ref{algorithm_stability_enforcement} 
may replace $\boldsymbol{K}_{p}$ in Eq.~\eqref{pd_grav} to ensure stability assuming the step size is sufficiently small, thus completing the formulation of our proposed adaptive controller in a manner that ensures stability.

\section{Experiments} \label{sec: experiments}
The proposed adaptive PD controller is implemented using the `ros\_control' module~\cite{ros_control} on a torque-controlled TALOS humanoid robot from PAL Robotics~\cite{talos_robot}.
It is then validated through simulation in Gazebo~\cite{talos_gazebo} as well as hardware experiments by comparing the performance of the proposed PD controller to that of a traditional non-adaptive PD controller.

Tab.~\ref{tab:controller parameters} lists the controller parameters used in the simulation and hardware experiments described in the following sections. The traditional controller uses $K_{po}$ and $K_{do}$ as gains. 
Four experiments are presented in this paper, validating the design objectives set for the controller in Sec.~\ref{sec: introduction}. Three experiments performed in simulation consist in tracking a sinusoidal trajectory, tracking a step input position, and response to an external force. The hardware experiment demonstrates the controller's response to an external force.
\begin{table}[t]
    \centering
    \caption{Controller Parameter Values Used In Simulation and Hardware Experiments}
    \begin{tabular}{ p{1.5cm} p{1.5cm} p{1.5cm} p{1.5cm} }
    \hline
    \textbf{Parameter}    & \textbf{Units}     & \textbf{Simulation}    & \textbf{Hardware}   \\ 
    \hline
    
    $\boldsymbol{K}_{po}$ & $\text{N~m~}\text{rad}^{-1}$ &  $10 \cdot \mathbf{1}_n$ &  $5 \cdot \mathbf{1}_n$ \\
    $\boldsymbol{K}_{do}$ & $\text{N~m~s~}\text{rad}^{-1}$ & $6 \cdot \mathbf{1}_n$ &  $ 0 \cdot \mathbf{1}_n$ \\
    $\boldsymbol{K}_{d_{max}}$ & $\text{N~m~s~}\text{rad}^{-1}$ & $20 \cdot \mathbf{1}_n$ &  $0.5 \cdot \mathbf{1}_n$ \\
    $\mathcal{L}_{lim}$ & J & $1.5$ & $0.15$ \\
    $\alpha$ & -- & $10$ & $3$\\
    $\beta$ & J & $1$ & $0.13$ \\
                
    \hline
    \end{tabular}
    \label{tab:controller parameters}
    \\ \vspace{0.1cm} $\mathbf{1}_n$ denotes the 1-vector of size $n$
\end{table}
\begin{table}[!t]
    \centering
    \caption{Reference Sinusoid Parameters for Simulation Experiment 1}
    \begin{tabular}{ p{1cm} p{2cm} p{2cm} p{2cm} }
    \hline
    \textbf{Joint}    & \textbf{Amplitude }     & \textbf{Frequency } 
    & \textbf{Offset}   \\ 
     & $A$ (rad) & $f$ (s$^{-1}$) &  $q_0$ (rad) \\
    \hline
    
    1 & 0.25 &  $1 /15$ &  0.5 \\
    2 & 0.5 &  $1/10$ &  -1 \\
    3 & 0.25 &  $1/8$ &  0 \\
    4 & 0.5 &  $1/7$ &  -1 \\
                
    \hline
    \end{tabular}
    \label{tab:sinusoid parameters}
    \\ \vspace{0.0cm} The reference sinusoid is generated as $q_r(t) = q_0 + A \sin{(2\pi f t})$ 
\end{table}

\begin{figure*}[!t]
    \centering
    \def\twidth{0.1}
    \subfloat[Traditional PD Controller Response]{\includegraphics[width=3.4in]{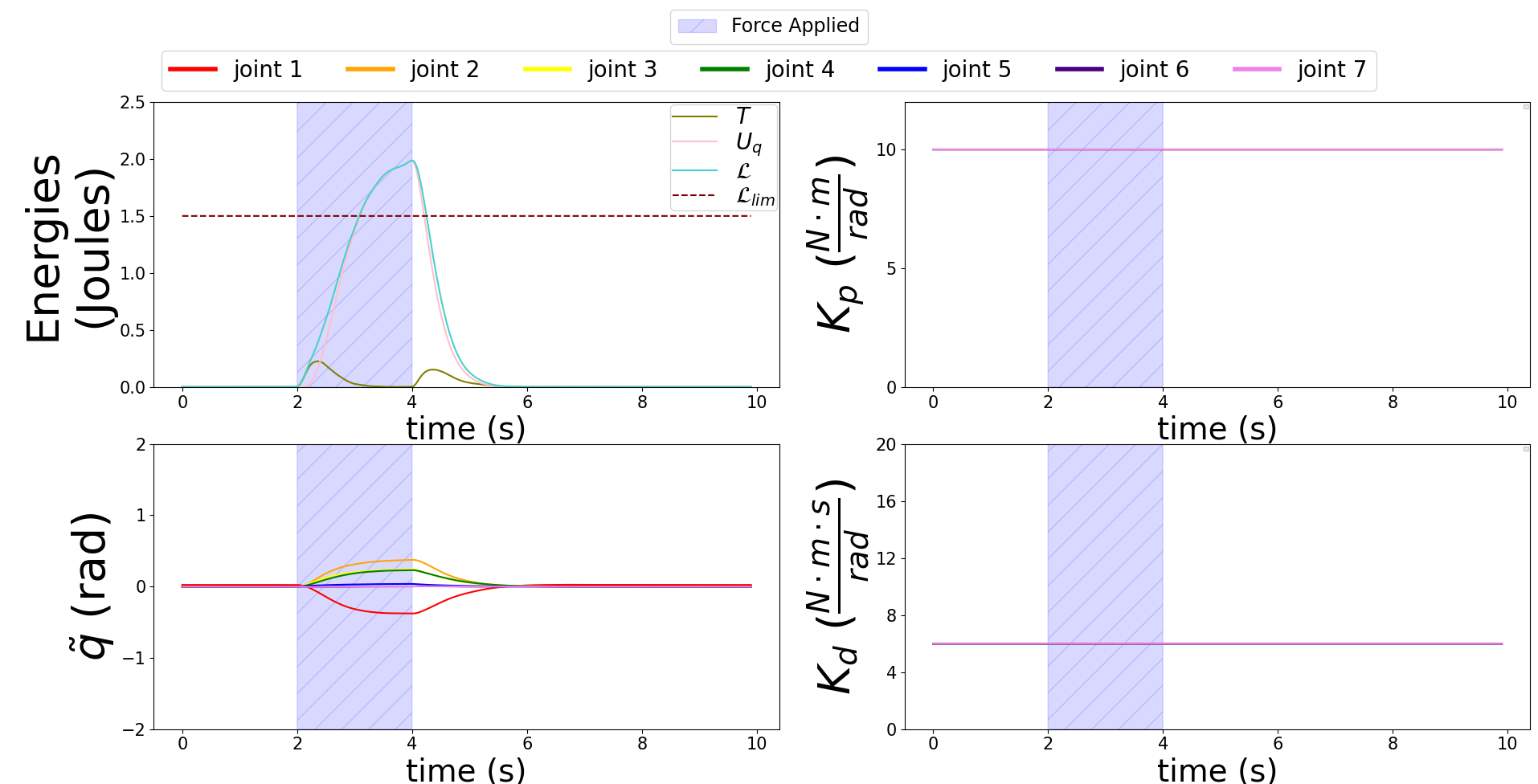}
    \label{fig_push_simple}}
    \hfil
    \subfloat[Adaptive PD Controller Response]{\includegraphics[width=3.4in]{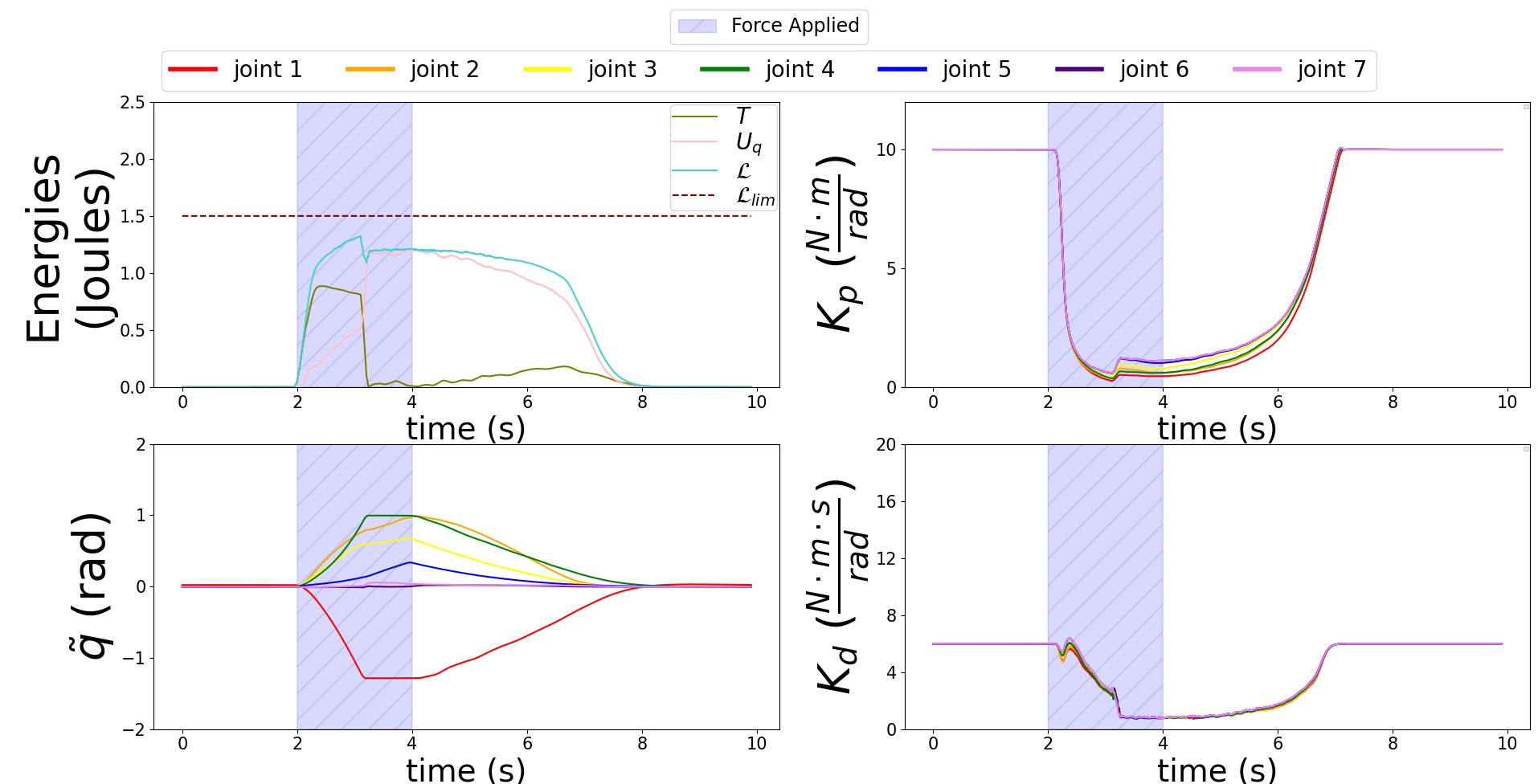}
    \label{fig_push_adaptive}}
    \caption{Controller responses in simulation for an external wrench $\boldsymbol{\mathcal{F}} = \begin{bmatrix}
        -10\text{ N} & 15\text{ N} & 10\text{ N} & 0\text{ Nm} & 0\text{ Nm} & 0.5\text{ Nm}
    \end{bmatrix}^\top$ applied to link 5 for a duration of $2~\text{s}$ while tracking a reference position $\boldsymbol{q}_r = \begin{bmatrix}
        0.5 & -1 & 0 & -1 & 0 & 0 & 0
    \end{bmatrix}^\top~\text{rad}$ for \protect\subref{fig_push_simple} the traditional 
    PD controller and \protect\subref{fig_push_adaptive} the proposed adaptive PD controller. 
    The blue shaded region in each plot indicates the time period during which the wrench is applied.
    }
    \label{fig_push}
\end{figure*}
\begin{figure*}[!t]
    \centering
    \def\twidth{0.1}
    \subfloat[Traditional PD Controller Response]{\includegraphics[width=3.4in]{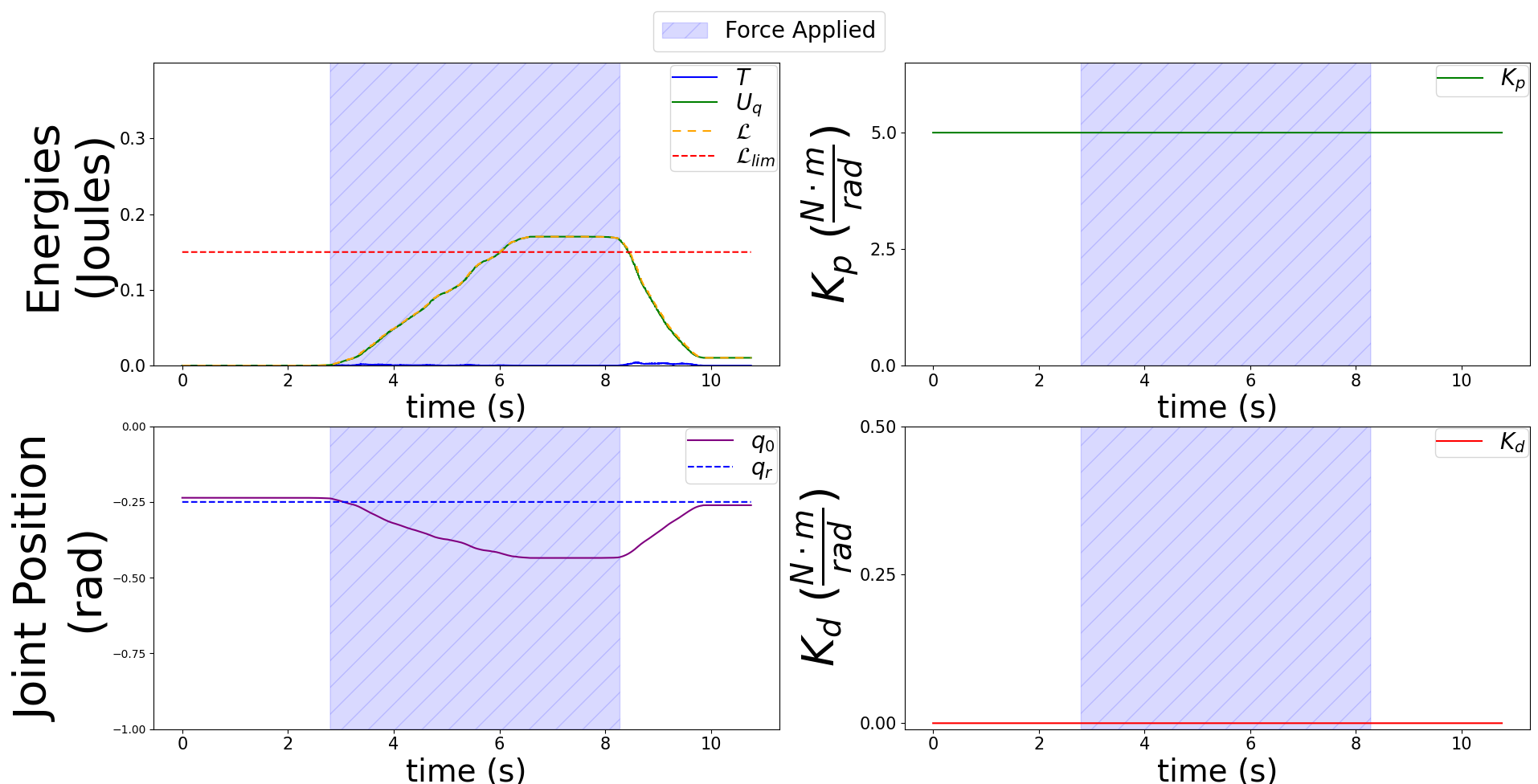}
    \label{fig_actual_push_simple}}
    \hfil
    \subfloat[Adaptive PD Controller Response]{\includegraphics[width=3.4in]{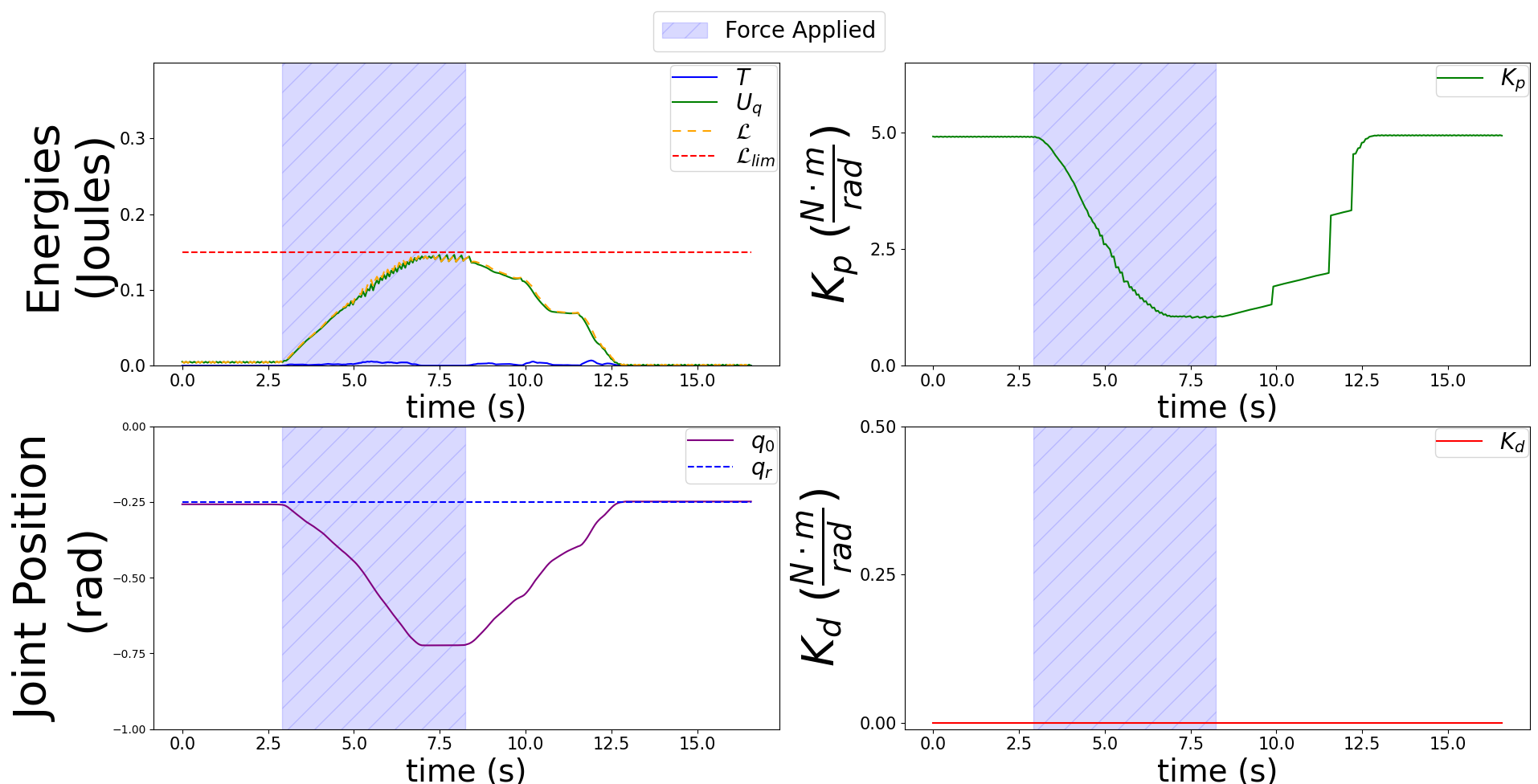}
    \label{fig_actual_push_adaptive}}
    \caption{Controller responses on hardware to an external force applied to link 1 of the robot's right arm while tracking a reference position for \protect\subref{fig_actual_push_simple} the traditional non-adaptive PD controller and \protect\subref{fig_actual_push_adaptive} the proposed adaptive PD controller. The energy and its components are shown for the entire arm. The PD gains and the position are shown only for joint 1, since that is the joint most directly affected by the applied force. The blue shaded region in each plot indicates the time period when the external force is applied.}
    \label{fig_actual_push}
\end{figure*}
\subsection{Simulation Experiment 1: Sinusoidal trajectory tracking} 
This experiment aims to showcase the controller performance when tracking a reference trajectory over time, for which the total energy of the robot remains far below the limit. 
Given sinusoidal reference trajectories for the first four joints of the robot's right arm, the tracking performance for the 7 joints of the arm obtained with the traditional PD controller is compared to that of our proposed adaptive PD controller. The parameters for the reference sinusoids are given in Tab.~\ref{tab:sinusoid parameters}, and results are shown in Fig.~\ref{fig_sine}. When tracking the sinusoidal trajectory, the total system energy remains low, resulting in only slight PD gain adaptation. The performance is consequently very similar for both controllers, as the adaptive controller succeeds in tracking the given trajectories. 
This experiment demonstrates that the gains remain close to nominally tuned values when the system energy is low.

\subsection{Simulation Experiment 2: Step Response}
In this experiment, all seven joints of the right arm of the robot are subjected to a step input: 
the reference position is changed instantly from an initial $\boldsymbol{q}_r = \boldsymbol{0}_7$ to a new reference position $\boldsymbol{q}_r = \begin{bmatrix}
    0.5 &  -1 & 0.2 & -1 & -0.5 & 0.5 & 0.2
\end{bmatrix}^\top~\text{rad}$.
The goal of this experiment is to validate the proposed controller's energy-limiting capabilities. We specifically expect the potential energy to be the main source of energy during this experiment. 
The results of this experiment are shown in Fig.~\ref{fig_tracking}. In Fig.~\ref{fig_tracking_simple}, a clear violation of the energy limit is seen for the traditional PD controller. The joints move quickly to the reference position, taking $2.09~\text{s}$ for all joints to settle within $0.05~\text{rad}$ of the goal position. On the other hand, Fig.~\ref{fig_tracking_adaptive} shows that our proposed adaptive PD controller succeeds in limiting the total energy of the controller below the limit by adapting the controller gains---the reduction in gains maintains the energy under the limit, and the gains gradually increase as the position error decreases. The tracking performance is also much more gradual, with the settling time being $3.21~\text{s}$ for the adaptive PD controller. These results validate the proposed controller's energy-limiting capabilities. While tracking performance may have been traded off, safer behaviour from a pHRI perspective is observed.

\subsection{Simulation Experiment 3: External Wrench Response } 
This experiment involves applying an external wrench to the robot's right arm, and monitoring the controller response to the push. This experiment aims to test the controller's compliance in response to external forces in addition to energy-limitation capabilities. A wrench $\boldsymbol{\mathcal{F}} = \begin{bmatrix}
    -10\text{ N} & 15\text{ N} & 10\text{ N} & 0\text{ Nm} &0\text{ Nm} & 0.5\text{ Nm}
\end{bmatrix}^\top$ is applied on link 5 (corresponding to the forearm) for a duration of $2~\text{s}$, while tracking a reference position $\boldsymbol{q}_r = \begin{bmatrix}
    0.5 & -1 & 0 & -1 & 0 & 0 & 0
\end{bmatrix}^\top~\text{rad}$. The result of this experiment is shown in Fig.~\ref{fig_push}. 
Fig. \ref{fig_push_simple} shows that in the traditional PD controller case, a violation of the energy limit is observed as a result of the external wrench. The maximum deflection measured across all joints is $0.375~\text{rad}$, meaning relatively stiff behaviour is observed. After the wrench is removed, $1.19~\text{s}$ are needed for all joints to settle within $0.05~\text{rad}$ of the reference position. Fig.~\ref{fig_push_adaptive} shows the results for the proposed adaptive PD controller. The controller succeeds in limiting the system's energy below the limit by adapting the gains. The proportional gains decrease to lower the system's potential energy, whereas the derivative gains initially stay high to oppose the high kinetic energy, before tapering off as the system's kinetic energy decreases. In this case, the robot's behaviour is much softer, as evidenced by a maximum deflection of $0.997~\text{rad}$---the arm acts compliantly as a direct consequence of limiting system energy. After release, the joints of the arm take $3.73~\text{s}$ to settle within $0.05~\text{rad}$ of $\boldsymbol{q}_r$. Once again, the adaptive controller response is much more gradual: tracking performance is sacrificed for safer pHRI behaviour. This experiment showcases the fact that energy-limitation naturally leads to compliant behaviour, while also demonstrating the controller's energy-limitation feature.
\subsection{Hardware Experiment: Response To An External Force}
\begin{figure}[!t]
\centering
\includegraphics[width=3in]{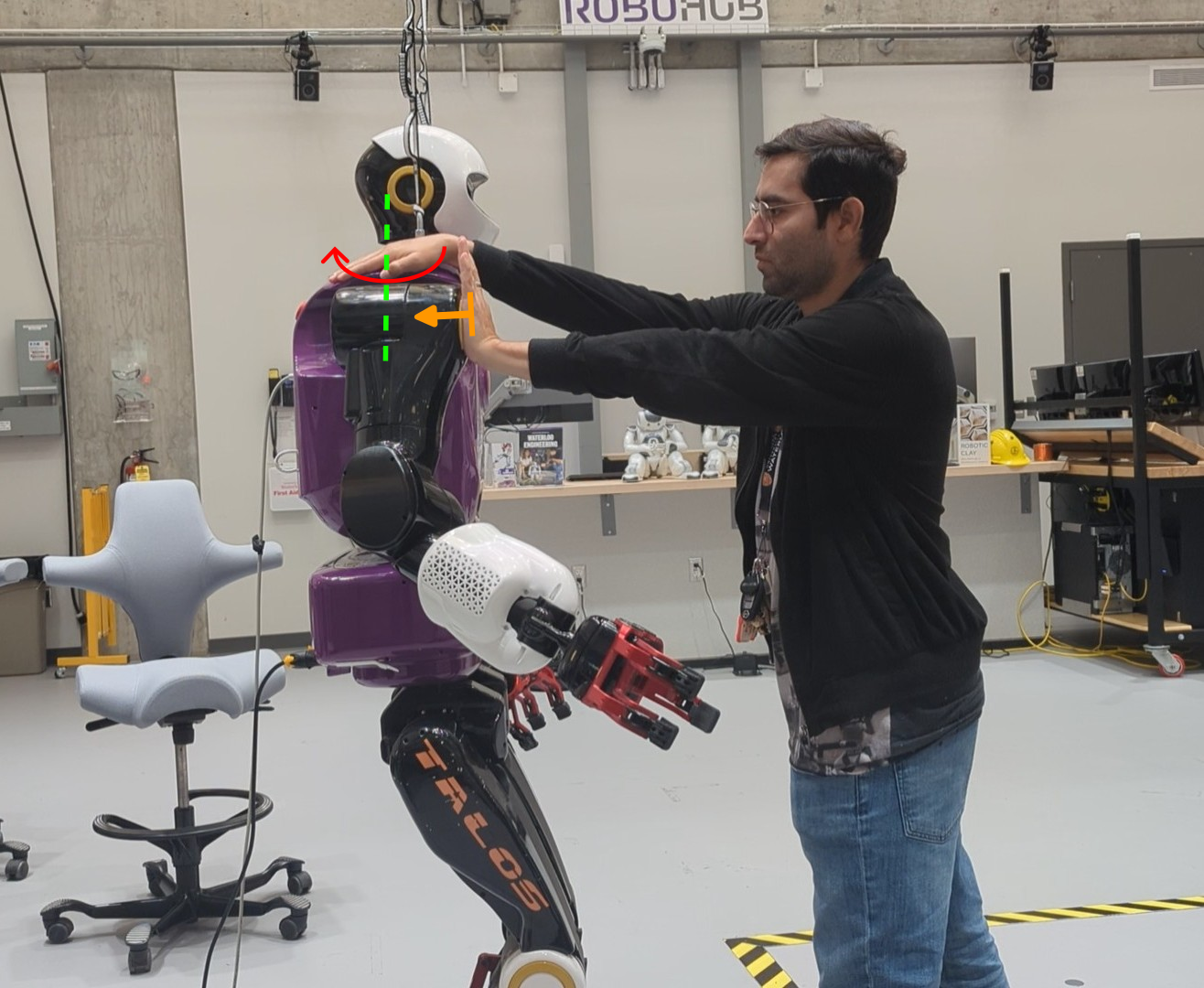}
\caption{The hardware experimental setup with the TALOS. The orange arrow denotes the direction of the applied force. The green dotted line denotes the axis of rotation of joint 1, whereas the red arrow shows the resulting joint rotation.}
\label{talos_setup}
\end{figure}

For this experiment, while the right arm of the TALOS robot tracks a reference position $\boldsymbol{q}_r = \begin{bmatrix}
    -0.25 & \boldsymbol{0}_6
\end{bmatrix}^\top~\text{rad}$, an external force is applied to the arm via pHRI at link 1 (corresponding to the shoulder), in the negative motion direction of joint 1. This setup is shown in Fig.~\ref{talos_setup}. The results of this interaction are shown in Fig. \ref{fig_actual_push} for both a traditional PD controller, and the proposed adaptive PD controller. For the traditional PD controller case shown in Fig. \ref{fig_actual_push_simple}, we observe a violation of the specified energy limit by the robot's total energy as a maximum value of $0.17~\text{J}$ is measured. The arm behaves stiffly in this case, with the joint deflection measured as $0.185~\text{rad}$. Once released, the arm takes $1.52~\text{s}$ to settle within $0.05~\text{rad}$ of the reference position. On the other hand, as shown in Fig. \ref{fig_actual_push_adaptive}, the adaptive PD controller succeeds in limiting the robot's total energy under the given limit. As the robot's arm is pushed, the adaptive PD gains slowly decrease, leading to reduced potential energy and reduced stiffness. The adaptive PD controller lends itself to softer, more compliant behaviour as well, since a greater deviation of $0.47~\text{rad}$ in the joint position is seen in the adaptive case despite lower total energy. The robot's tracking response is also more gradual, with the arm taking $4.30~\text{s}$ after the release of the force to settle back within $0.05~\text{rad}$ of $\boldsymbol{q}_r$.

\section{Discussion} \label{sec: discussion} 
The proposed approach can successfully limit the total energy of the robot under a given limit when tracking a reference position or trajectory, while also appropriately opposing user-added energy. The controller is simple to implement as it is an adaptive PD controller, and has the advantage of behaving as a traditional PD controller when far from the energy limit. The nominal PD gains may thus be tuned using any conventional tuning method. 

Like many other energy-based control schemes, the proposed controller requires accurate dynamic modelling data to be available. Energy-limitation is also not a necessary guarantee of safety, since contact area and consequent applied pressure during pHRI is an additional quantity to be considered---energy limitation does not necessarily eliminate puncture risks to humans, depending on the specific mechanical shape and properties of the robot. Energy limitation may also indirectly limit the maximum load a robot can carry.

Specifically for our proposed control scheme, appropriate selection of $\alpha$ and $\beta$ is necessary for successful energy limitation. The robot's resistance to external forces is also dependent on the tunable parameter $\boldsymbol{K}_{d_{max}}$: wrenches large enough to overcome this additional damping may lead to violations of the energy limit. This parameter is left tunable, since very high resistance to external pushes may not necessarily be desirable in pHRI \cite{iso_1,energy_budget_1,robot_safety_1,robot_safety_2}. The approach also supposes the existence of an algorithm that defines appropriate energy limits in different interaction scenarios to ensure safety.

\section{Conclusion}\label{sec: conclusion}
In this paper, we present an adaptive PD control scheme that can limit a robot's total energy to be under any given limit, to adhere to the ISO/TS 15066 standards for safe pHRI. The controller has provisions to address the high kinetic and potential energy of a system. A Lyapunov analysis of the proposed controller yielded a stability condition for the controller. Experimental testing of the controller both in simulation and on hardware verified the expected controller behaviour and characteristics. Future work will focus on further hardware experimental validation of the controller in a wider variety of collision scenarios, extending to the entire robot. Inertia-based energy budget allocation to joints may be explored. The conservation of PD parameters such as the natural frequency and the damping ratio as the controller gains change may be investigated. We also hope to examine the effect of friction compensation alongside the controller in hardware. Finally, the stability proof may be expanded upon to explicitly incorporate changing reference positions, although this may require further theoretical development.

\bibliographystyle{ieeetr}
\bibliography{references}

\begin{thebibliography}{10}

\bibitem{pHRI_1}
A.~De~Santis, B.~Siciliano, A.~Luca, and A.~Bicchi, ``An atlas of physical human-robot interaction,'' {\em Mechanism and Machine Theory}, vol.~43, pp.~253--270, 03 2008.

\bibitem{farajtabar_hri}
M.~Farajtabar and M.~Charbonneau, ``The path towards contact-based physical human–robot interaction,'' {\em Robotics and Autonomous Systems}, vol.~182, p.~104829, 2024.

\bibitem{compliance_1}
A.~Calanca, R.~Muradore, and P.~Fiorini, ``A review of algorithms for compliant control of stiff and fixed-compliance robots,'' {\em IEEE/ASME Transactions on Mechatronics}, vol.~21, no.~2, pp.~613--624, 2016.

\bibitem{compliance_2}
A.~Q. Keemink, H.~van~der Kooij, and A.~H. Stienen, ``Admittance control for physical human–robot interaction,'' {\em The International Journal of Robotics Research}, vol.~37, no.~11, pp.~1421--1444, 2018.

\bibitem{compliance_3}
M.~Schumacher, J.~Wojtusch, P.~Beckerle, and O.~von Stryk, ``An introductory review of active compliant control,'' {\em Robot. Auton. Syst.}, vol.~119, p.~185–200, Sept. 2019.

\bibitem{energy_budget_1}
J.~Lachner, F.~Allmendinger, N.~Hogan, and S.~Stramigioli, ``Energy budgets for coordinate invariant robot control in physical human–robot interaction,'' {\em The International Journal of Robotics Research}, vol.~40, 05 2021.

\bibitem{iso_ts_15066_2016}
{International Organization for Standardization}, ``{ISO/TS 15066: Robots and robotic devices - Collaborative robots},'' 2016.
\newblock https://www.iso.org.

\bibitem{kellypd1997}
R.~Kelly, ``{PD} {Control} with {Desired} {Gravity} {Compensation} of {Robotic} {Manipulators}: {A} {Review},'' {\em The International Journal of Robotics Research}, vol.~16, pp.~660--672, Oct. 1997.
\newblock Publisher: SAGE Publications Ltd STM.

\bibitem{contact_1}
A.~De~Luca, A.~Albu-Schaffer, S.~Haddadin, and G.~Hirzinger, ``Collision detection and safe reaction with the dlr-iii lightweight manipulator arm,'' in {\em 2006 IEEE/RSJ International Conference on Intelligent Robots and Systems}, pp.~1623--1630, 2006.

\bibitem{iso_cbf}
F.~Ferraguti, M.~Bertuletti, C.~T. Landi, M.~Bonfè, C.~Fantuzzi, and C.~Secchi, ``A control barrier function approach for maximizing performance while fulfilling to iso/ts 15066 regulations,'' {\em IEEE Robotics and Automation Letters}, vol.~5, no.~4, pp.~5921--5928, 2020.

\bibitem{pfl_1}
A.~Golshani, A.~Kouhkord, A.~Ghanbarzadeh, and E.~Najafi, ``Control design for safe human-robot collaboration based on iso/ts 15066 with power and force limit,'' in {\em 2023 11th RSI International Conference on Robotics and Mechatronics (ICRoM)}, pp.~279--284, 2023.

\bibitem{pfl_2}
P.~Aivaliotis, S.~Aivaliotis, C.~Gkournelos, K.~Kokkalis, G.~Michalos, and S.~Makris, ``Power and force limiting on industrial robots for human-robot collaboration,'' {\em Robotics and Computer-Integrated Manufacturing}, vol.~59, pp.~346--360, 2019.

\bibitem{force_estimation_1}
Y.~Lu, Y.~Shen, and Z.~Chungang, ``External force estimation for industrial robots using configuration optimization,'' {\em Automatika}, vol.~64, pp.~1--24, 01 2023.

\bibitem{force_estimation_2}
M.~Van~Damme, P.~Beyl, B.~Vanderborght, V.~Grosu, R.~Van~Ham, I.~Vanderniepen, A.~Matthys, and D.~Lefeber, ``Estimating robot end-effector force from noisy actuator torque measurements,'' in {\em 2011 IEEE International Conference on Robotics and Automation}, pp.~1108--1113, 2011.

\bibitem{collision_reaction_1}
S.~Haddadin, A.~Albu-Schaffer, A.~De~Luca, and G.~Hirzinger, ``Collision detection and reaction: A contribution to safe physical human-robot interaction,'' in {\em 2008 IEEE/RSJ International Conference on Intelligent Robots and Systems}, pp.~3356--3363, 2008.

\bibitem{iso_1}
F.~Benzi, F.~Ferraguti, and C.~Secchi, ``Energy tank-based control framework for satisfying the iso/ts 15066 constraint,'' {\em IFAC-PapersOnLine}, vol.~56, no.~2, pp.~1288--1293, 2023.
\newblock 22nd IFAC World Congress.

\bibitem{energy_transfer_1}
J.~Liu, Y.~Yamada, and Y.~Akiyama, ``Calculating the supplied energy for physical human-robot interaction,'' in {\em 2021 IEEE International Conference on Intelligence and Safety for Robotics (ISR)}, pp.~157--160, 2021.

\bibitem{force_direction_1}
O.~Khatib, ``Inertial properties in robotic manipulation: An object-level framework,'' {\em The International Journal of Robotics Research}, vol.~14, no.~1, pp.~19--36, 1995.

\bibitem{energy_tank_1}
F.~Ferraguti, C.~Secchi, and C.~Fantuzzi, ``A tank-based approach to impedance control with variable stiffness,'' in {\em 2013 IEEE International Conference on Robotics and Automation}, pp.~4948--4953, 2013.

\bibitem{meguenani2017energy}
A.~Meguenani, V.~Padois, J.~Da~Silva, A.~Hoarau, and P.~Bidaud, ``Energy based control for safe human-robot physical interaction,'' in {\em 2016 International Symposium on Experimental Robotics}, pp.~809--818, Springer, 2017.

\bibitem{laffranchi2009safe}
M.~Laffranchi, N.~G. Tsagarakis, and D.~G. Caldwell, ``Safe human robot interaction via energy regulation control,'' in {\em 2009 IEEE/RSJ International Conference on Intelligent Robots and Systems}, pp.~35--41, IEEE, 2009.

\bibitem{hjorth2020energy}
S.~Hjorth, J.~Lachner, S.~Stramigioli, O.~Madsen, and D.~Chrysostomou, ``An energy-based approach for the integration of collaborative redundant robots in restricted work environments,'' in {\em 2020 IEEE/RSJ International Conference on Intelligent Robots and Systems (IROS)}, pp.~7152--7158, IEEE, 2020.

\bibitem{choi2024multi}
S.~Choi, S.~Ha, and W.~Kim, ``A multi-task energy-aware impedance controller for enhanced safety in physical human-robot interaction,'' {\em IEEE Robotics and Automation Letters}, 2024.

\bibitem{groothuis2018stabilityenergy}
S.~S. Groothuis, G.~A. Folkertsma, and S.~Stramigioli, ``A general approach to achieving stability and safe behavior in distributed robotic architectures,'' {\em Frontiers in Robotics and AI}, vol.~5, 2018.

\bibitem{folkertsma2018safetystability}
G.~A. Folkertsma, S.~S. Groothuis, and S.~Stramigioli, ``Safety and guaranteed stability through embedded energy-aware actuators,'' in {\em 2018 IEEE International Conference on Robotics and Automation (ICRA)}, pp.~2902--2908, 2018.

\bibitem{zanella2024optimal}
R.~Zanella, A.~Macchelli, and S.~Stramigioli, ``Learning the optimal energy-based control strategy for port-hamiltonian systems,'' {\em IFAC-PapersOnLine}, vol.~58, no.~6, pp.~208--213, 2024.
\newblock 8th IFAC Workshop on Lagrangian and Hamiltonian Methods for Nonlinear Control LHMNC 2024.

\bibitem{califano2024effectcontrolbarrierfunctions}
F.~Califano, R.~Zanella, A.~Macchelli, and S.~Stramigioli, ``The effect of control barrier functions on energy transfers in controlled physical systems,'' 2024.

\bibitem{hjorth2023energystability}
S.~Hjorth, E.~Lamon, D.~Chrysostomou, and A.~Ajoudani, ``Design of an energy-aware cartesian impedance controller for collaborative disassembly,'' in {\em 2023 IEEE International Conference on Robotics and Automation (ICRA)}, pp.~7483--7489, 2023.

\bibitem{kelly2005control}
R.~Kelly, V.~Davila, and J.~Perez, {\em Control of Robot Manipulators in Joint Space}.
\newblock Advanced Textbooks in Control and Signal Processing, Springer London, 2005.

\bibitem{spong2005robot}
M.~Spong, S.~Hutchinson, and M.~Vidyasagar, {\em Robot Modeling and Control}.
\newblock Wiley, 2005.

\bibitem{ott_cartesian}
C.~Ott, ``Cartesian impedance control of redundant and flexible-joint robots,'' in {\em Springer Tracts in Advanced Robotics}, 2008.

\bibitem{passivity_2}
M.~Lemmon, ``Passivity based control,'' 2017.
\newblock https://www3.nd.edu/~lemmon/courses/ee580/lectures/chapter10.pdf, Last accessed on 2024-10-15.

\bibitem{pd_stability_q_d}
S.~Kawamura, F.~Miyazaki, and S.~Arimoto, ``Is a local linear pd feedback control law effective for trajectory tracking of robot motion?,'' in {\em Proceedings. 1988 IEEE International Conference on Robotics and Automation}, pp.~1335--1340 vol.3, 1988.

\bibitem{hi-per}
J.~Chen and P.~I. Ro, ``Human intention-oriented variable admittance control with power envelope regulation in physical human-robot interaction,'' {\em Mechatronics}, vol.~84, p.~102802, 2022.

\bibitem{ros_control}
S.~Chitta, E.~Marder-Eppstein, W.~Meeussen, V.~Pradeep, A.~Rodr{\'i}guez~Tsouroukdissian, J.~Bohren, D.~Coleman, B.~Magyar, G.~Raiola, M.~L{\"u}dtke, and E.~Fern{\'a}ndez~Perdomo, ``ros\_control: A generic and simple control framework for ros,'' {\em The Journal of Open Source Software}, 2017.

\bibitem{talos_robot}
{PAL Robotics}, ``{The TALOS Robot}.''
\newblock https://pal-robotics.com/robot/talos/, Last accessed on 2024-11-29.

\bibitem{talos_gazebo}
{Víctor López, Sai Kishor Kothakota}, ``{Installing TALOS Simulation}.''
\newblock https://wiki.ros.org/Robots/TALOS/Tutorials/Installation/Simulation, Last accessed on 2024-11-10.

\bibitem{robot_safety_1}
A.~Pervez and J.~Ryu, ``Safe physical human robot interaction-past, present and future,'' {\em Journal of Mechanical Science and Technology}, vol.~22, pp.~469--483, 2008.

\bibitem{robot_safety_2}
A.~Bicchi, M.~A. Peshkin, and J.~E. Colgate, {\em Safety for Physical Human--Robot Interaction}, pp.~1335--1348.
\newblock Berlin, Heidelberg: Springer Berlin Heidelberg, 2008.

\end{thebibliography}

\end{document}